\def\year{2020}\relax
\documentclass[letterpaper]{article} 
\usepackage{aaai20}  
\usepackage{times}  
\usepackage{helvet} 
\usepackage{courier}  
\usepackage[hyphens]{url}  
\usepackage{graphicx} 
\usepackage{graphicx}  
\usepackage{enumitem}
\usepackage{soul}
\usepackage{url}
\usepackage[utf8]{inputenc}
\usepackage{graphicx}
\usepackage{amsmath}
\usepackage{booktabs}
\usepackage{algorithm}
\usepackage{algorithmic}
\usepackage{subfigure}
\RequirePackage[textsize=scriptsize]{todonotes}
\usepackage{multicol,lipsum}

\usepackage{amsmath}
\usepackage{makecell}
\usepackage{url}
\usepackage{todonotes}
\usepackage[normalem]{ulem} 
\usepackage{soul} 

\usepackage{xcolor}
\usepackage{graphicx}
\usepackage{soul}
\usepackage[utf8]{inputenc}
\usepackage{amsmath}
\usepackage{bm}
\usepackage{pbox}
\usepackage{multirow}
\newcommand{\citet}[1]{\citeauthor{#1} \shortcite{#1}}
\newcommand{\citep}{\cite}

\newcommand{\vek}[1]{{\mathbf {#1}}}
\newcommand{\vx}{{\vek{x}}}
\newcommand{\vw}{{\vek{w}}}
\newcommand{\vs}{{\vek{s}}}
\newcommand{\vtau}{{\vek{\tau}}}
\newcommand{\cY}{{\mbox{$\mathcal{Y}$}}}
\newcommand{\cX}{{\mbox{$\mathcal{X}$}}}

\newcommand{\msg}{{\mbox{M}}}
\newcommand{\model}{\mbox{CAGE}\ }
\newcommand{\modelNS}{\mbox{CAGE}}

\newcommand{\modelnoconst}{\mbox{CAGE}$_{-\text{G}}$}
\newcommand{\modeldep}{\mbox{CAGE}$_{\text{dataG}}$}
\newcommand{\modeldepnocont}{\mbox{CAGE}$_{\text{-C,dataG}}$}
\newcommand{\modelnoconstnocont}{\mbox{CAGE}$_{-\text{C}-\text{G}}$}
\newcommand{\modelpenalty}{\mbox{CAGE}$_{-\text{C}-\text{G}+-\text{P}}$}
\newcommand{\modelnocont}{\mbox{CAGE}$_{-\text{C}}$}

\begin{document}
\title{Data Programming using Continuous and Quality-Guided Labeling Functions}

 \author{Oishik Chatterjee \\
   Department of CSE\\
   IIT Bombay, India \\
   \qquad \qquad {\tt oishik@cse.iitb.ac.in}
   \And 
  Ganesh Ramakrishnan\\
  Department of CSE\\ IIT Bombay, India\\
  \qquad \qquad {\tt  ganesh@cse.iitb.ac.in}
   \And
   Sunita Sarawagi \\
   Department of CSE\\
   IIT Bombay, India \\
   \qquad{\tt sunita@iitb.ac.in} \\\qquad
  }

\maketitle
\begin{abstract}
Scarcity of labeled data is a bottleneck for supervised learning models. A  paradigm that has evolved for dealing with this problem is data programming.
An existing data programming paradigm allows human supervision to be provided as a set of discrete labeling functions (LF)  that output possibly noisy labels to input instances and a generative model for consolidating the weak labels. 
We enhance and generalize this paradigm by supporting functions that output a continuous score (instead of a hard label) that noisily correlates with labels. We show across five applications that continuous LFs are more natural to program and lead to improved recall. 
We also show that accuracy of existing generative models is unstable with respect to initialization, training epochs, and learning rates.  We give control to the data programmer to guide the training process by providing intuitive quality guides with each LF.  We propose an elegant method of incorporating these guides into the generative model.  Our overall method, called \mbox{CAGE}, makes the data programming paradigm more reliable than other tricks based on initialization, sign-penalties, or soft-accuracy constraints.         
\end{abstract}


\section{Introduction}
\label{Intro}
Modern machine learning systems require large amounts of labelled data.
For many applications, such labelled data is created by getting humans  to explicitly label each training example. A problem of perpetual interest in machine learning is reducing the tedium of such human supervision via techniques like active learning, crowd-labeling, distant supervision, and semi-supervised learning.  A limitation of all these methods is that supervision is restricted to the level of individual examples.

A recently proposed~\citep{snorkel_nips2016}  paradigm is that of Data Programming.  In this paradigm, humans provide several labeling functions  written in any high-level programming language.  Each labeling function (LF) takes as input, an example and either attaches a label to it or backs off. 
We illustrate such LFs on one of the five tasks that we experimented with, {\it viz.}, that of labeling mention of  a pair of people names  in a sentence as defining the {\it spouse} relation or not.
The users construct heuristic patterns as LFs for identifying spouse relation in a sentence containing an entity pair $\left( E_1, E_2\right)$. 
A LF can assign +1 to indicate that the {\it spouse} relation is true for the candidate pair ($E_1$,$E_2$), -1 to mean that no {\it spouse} relation, and 0 to mean that the LF in unable to assert anything for this example.  Specifically for the {\it spouse} relation extraction task,  Table~\ref{tab:sample} lists six LFs. 


\begin{table*}[htb]
\small
\centering 
\begin{tabular}{|l|l|} 
\hline
\multicolumn{2}{|l|}
{{\bf SpouseDict} = \{'spouse', 'married', wife', 'husband', 'ex-wife', 'ex-husband'\}} \\
\multicolumn{2}{|l|}
{{\bf FamilyDict}  = \{'father', 'mother', 'sister', 'brother', 'son', 'daughter','grandfather', 'grandmother', 'uncle', 'aunt','cousin' \}$\bigotimes$\{+'',+'-in-law'\} } \\
\multicolumn{2}{|l|}{{\bf OtherDict}  =  \{'boyfriend', 'girlfriend' 'boss', 'employee', 'secretary', 'co-worker'\} }\\ 
\multicolumn{2}{|l|}{{\bf SeedSet}  =  \{('Barack Obama', 'Michelle Obama
 '), ('Jon Bon Jovi','Dorothea Hurley'),('Ron Howard','Cheryl Howard'),.....\} }\\ 
\hline
{\bf Id} &  \textbf{Description} \\
\hline
{\bf LF1} &  If some word in {\bf SpouseDict} is present between $E_1$ and $E_2$ or within 2 words of either, return 1 else return 0 \\
\hline
{\bf LF2} & If some word in {\bf FamilyDict} is present between $E_1$ and $E_2$, return -1 else return 0.\\ \hline
{\bf LF3} & If some word in {\bf OtherDict} is present between $E_1$ and $E_2$, return -1 else return 0.\\ \hline
{\bf LF4} & If both $E_1$ and $E_2$ occur in  {\bf SeedSet}, return 1 else return 0.\\ \hline
{\bf LF5}   & If the number of word tokens lying between $E_1$ and $E_2$ are less than 4, return 1 else return 0. \\ \hline
\end{tabular}
\caption{Discrete LFs based on dictionary lookups or thresholded distance for the {\it spouse} relationship extraction task}\label{tab:sample}
\end{table*}

In isolation, each LF may neither be always correct nor complete. LFs may also produce conflicting labels.
For the purpose of illustration, consider a text snippet `Michelle Obama is the mother of Malia and Sasha and the wife of Barack Obama'. For the candidate pair (`Michelle Obama', 'Barack Obama'),  LF1 and LF4 in Table~\ref{tab:sample}  assign a label 1 whereas LF2 assigns the label -1. 

\citet{snorkel_nips2016} presented a generative model for consensus on the noisy and conflicting labels assigned by the discrete LFs to determine probability of the correct labels. Labels thus obtained could be used for training any supervised model/classifier and evaluated on a  test set. 
%
%
In this paper, we present two significant extensions of the above data programming paradigm.

First, the user provided set of LFs might not be complete in their discrete forms. LF1 through LF3 in Table~\ref{tab:sample} that look for words in various hand-crafted dictionaries, may have incomplete dictionaries.  
\begin{table*}[htb]
\small
\centering 
\begin{tabular}{|l|l|l|} 
\hline
\hline
{\bf Id} & {\bf Class} &  \textbf{Description} \\
\hline
{\bf LF1} & +1 & $\max $ [cosine(word-vector($u$), word-vector($v$))-0.8$]_+$: $u \in $ {\bf SpouseDict} and $v \in$ \{words between $E_1, E_2$\}. \\
\hline
{\bf LF2} & -1 & $\max $ [cosine(word-vector($u$), word-vector($v$))-0.8$]_+$: $u \in $ {\bf FamilyDict} and $v \in$ \{words between $E_1, E_2$\}.  \\ \hline
{\bf LF3} & -1 & $\max $ [cosine(word-vector($u$), word-vector($v$))-0.8$]_+$: $u \in $ {\bf OtherDict} and $v \in$ \{words between $E_1, E_2$\}. \\ \hline
{\bf LF4} & -1 & $\max $ [0.2 -  Norm-Edit-Dist($E_1,E_2$, $u,v$)$]_+$: ~$(u,v), (v,u) \in $ {\bf SeedSet}. \\ \hline
{\bf LF5} & +1 & [1 - (number of word tokens between $E_1$ and $E_2$)/5.0$]_+$ \\ \hline
\end{tabular}
\caption{Continuous LFs corresponding to some of the discrete LFs in Table~\ref{tab:sample} for the {\it spouse} relationship extraction task}\label{tab:continuoussample}
\end{table*}
A more comprehensive alternative could be to design continuous valued LFs that return scores derived from soft match between words in the sentence and the dictionary.  As an example, for LF1 through LF3, the soft match could be obtained based on cosine similarity of pre-trained word embedding vectors~\citep{word2vec} of a word in the dictionary with a word in the sentence. 
This could enable an LF to provide a continuous class-specific score
to the model, instead of a hard class label  (when triggered).  In Table~\ref{tab:continuoussample}, we list a continuous LF corresponding to each LF from Table~\ref{tab:sample}.  Such  
continuous LFs can expand the scope of matching to semantically similar words beyond the pre-specified words in the dictionary.  For example: in the sentence {\tt 1) $<$Allison$>$, 27, and $<$Ricky$>$, 34, wed on Saturday surrounded by friends.}, the word 'wed' is semantically similar to 'married' and would be detected by our continuous LF but missed by the discrete ones in Table~\ref{tab:sample}.


More generally across applications, human experts are very often able to identify real-valued scores that correlate strongly with the label  but find it difficult to discretize that score into a hard label.  More examples of such scores include TF-IDF match with prototypical documents in a text-classification task, distance among entity pairs for a relation extraction task,  and confidence scores from hand-crafted classifiers. 
We extend existing generative models~\citep{snorkel_nips2016,snorkel_icml2017} to support continuous LFs. In addition to modeling the consensus distribution over all LFs (continuous and discrete), we model the distribution of scores for each continuous LF. 
In Section~\ref{sec:motivationExample} 
in the supplementary material, 
we illustrate through an example, why the model for continuous LFs is not a straightforward extension of the discrete counterpart.

Our second extension is designed to remove the
instability of existing generative training based on unsupervised likelihood.  Across several datasets we observed  that the accuracy obtained by the existing models was highly sensitive to initialization, training epochs, and learning rates.  In the absence of labeled validation set, we cannot depend on prevalent tricks like early stopping to stabilize training.  We give control to the data programmer to guide the training process by providing intuitive accuracy guides with each LF.
In the case that the labeler develops each LF after inspecting some examples, the labeler would naturally have a rough estimate of the 
fraction $q$ of examples that would be correctly labeled by the LF, amongst all examples that trigger the LF. Even otherwise, it might not be too difficult for the labeler to intuitively specify some value of $q$. 
We show that such a $q$ serves as a user-controlled quality guide that can effectively guide the training process. 
For the case of continuous LFs, we use  $q$
as a rough estimate of the mean score of the continuous LF whenever it triggers correctly. A quality guide of $q=0.9$ for LF1 would imply that when LF1 triggers correctly the average embedding score is 0.9. This is easier than choosing a hard threshold on the embedding score to convert it to a discrete LF.   We provide an elegant method of guiding our generative training with these user-provided accuracy estimates, and show how it surpasses simpler methods like sign penalty and data constraints.  Empirically, we show our method stabilizes unsupervised likelihood training even with very crude estimates of $q$.
We study stability issues of the existing model  with respect to training epochs and demonstrate that the  proposed model is naturally more stable. 
We refer to our overall approach as \mbox{CAGE}, which stands for {\bf \uline{ C}}ontinuous {\bf \uline{ A}}nd quality {\bf \uline{ G}}uided lab{\bf \uline{ E}}ling functions. In summary, \model makes the following contributions: \\
1) It enhances the expression power of data programming by supporting continuous labeling functions (generalizations of discrete LFs) as  the unit of supervision from humans.\\
2) It proposes a carefully parameterized graphical model that outperforms existing models even for discrete LFs, and permits easy incorporation of user priors for continuous LFs.\\
3) It extends the generative model  through quality guides, thereby increasing its stability and making it less sensitive to initialization. Its training is based on a principled method of regularizing the marginals of the joint model with user-provided accuracy guides.
\\
We present extensive experiments on five datasets, comparing various models for performance and stability, and present the significantly positive impact of \model.  We show that our method of incorporating user guides leads to more reliable training than obvious ideas like sign penalty and constraint based training.  


\section{Our Approach: \model}
\label{ourapp}
Let $\cX$ denote the space of input instances and $\cY=\{1,\ldots,K\}$ denote the space of labels and $P(\cX,\cY)$ denote their joint distribution. Our goal is to learn a model to associate a label $y$ with an example $\vx \in \cX$. Unlike standard supervised learning, we are not provided true labels of sampled instances during training.  Let the  sample of $m$ unlabeled instances be $\vx_1,\ldots,\vx_m$.  Instead of the true $y$'s we are provided a set of $n$ labeling functions (LFs)  $ \lambda_1, \lambda_2, \ldots \lambda_n $ such that each LF  $\lambda_j$ can be either discrete or continuous. Each LF $\lambda_j$ is attached with a class $k_j$ and on an instance $\vx_i$ outputs a discrete label $\tau_{ij}=k_j$ when triggered and  $\tau_{ij}=0$ when not triggered.  If $\lambda_j$ is continuous, it also outputs a score $s_{ij} \in (0,1)$. This is a form of weak supervision that implies that when a LF is triggered on an instance $\vx_i$, it is proposing that the true label $y$  should be $k_j$, and if continuous it is attaching a  confidence proportional to $s_{ij}$ with its labeling.  

But to reliably and accurately infer the true label $y$ from such weak supervision without any labeled data,  we need to exploit the assumption that $\tau_{ij}$ is positively correlated with $y$.  We allow the programmer of the LF to make this assumption explicit by attaching a guess on the fraction $q_j^t$ of triggering of the LF where the true $y$ agrees with $\tau_{ij}$.   We show in the experiments that crude guesses suffice.
Intuitively, this says that the user expects $q_j^t$ fraction of examples for which the LF value has been triggered to be correct.  Additionally, for continuous LF the programmer can specify the quality index  $q_j^c$ denoting the average score of $s_j$ when there is such agreement.

Our goal is to learn to infer the correct label by creating consensus among outputs of the LFs.
Thus, the model of \model imposes a joint distribution between the true label $y$ and the values $\tau_{ij}, s_{ij}$ returned by each LF $\lambda_j$ on any data sample $\vx_i$ drawn from the hidden distribution $P(\cX,\cY)$. 
\begin{equation}
\displaystyle P_{\theta,\pi}(y,\vtau_{i},\vs_i) = \frac{1}{Z_\theta} \prod_{j=1}^n \psi_\theta(\tau_{ij}, y) \left( \psi_\pi(\tau_{ij}, s_{ij}, y)\right)^{\text{cont}(\lambda_j)}
\label{eq:joint}
\end{equation}
where $\text{cont}(\lambda_j)$ is $1$ when $\lambda_j$ is a continuous LF and $0$ otherwise. And $\theta,\pi$ denote the parameters used in defining the potentials $psi_\theta, \psi_\pi$ coupling discrete and continuous variables respectively.
In this factorization of the joint distribution we make the natural assumption that each LF independently provides its supervision on the true label.  The main challenge now is designing the potentials coupling various random variables so that: (a) The parameters ($\theta,\pi$) can be trained reliably using unlabeled data alone. This partially implies that the number of parameters should be limited. (b) The model should be expressive enough to fit the joint distribution on the $\tau_j$ and $s_j$ variables across a variety  of datasets without relying on labeled validation dataset for model selection and hyper-parameter tuning. (c) Finally, the potentials should reflect the bias of the programmer on the quality in providing the true $y$.  We will show how without such control, it is easy to construct counter-examples where the standard likelihood-based training may fail miserably.

With these goals in mind, and after significant exploration we propose the following form of potentials.  For the discrete binary $\tau_{ij}$ variables, we chose these simple potentials:

\begin{equation}
    \psi_{\theta}(\tau_{ij},y) = 
\begin{cases}
    \exp(\theta_{jy})  & \text{if $\tau_{ij}\ne 0$,} \\
    1 & \text{otherwise.}
\end{cases}
\label{eq:decoupledthetas}
\end{equation}
Thus, for each LF we have $K$ parameters corresponding to each of the class labels.    An even simpler alternative would be to share the $\theta_{jy}$ across different $y$s as used in \cite{snorkel_icml2017} but that approach imposes undesirable restrictions on the distributions it can express.  We elaborate on that in Section~\ref{sec:compare}.

For the case of continuous LFs the task of designing the potential $\psi_{\pi}(s_{ij}, \tau_{ij}, y)$  that is trainable with unlabeled data and captures user bias well turned out to be significantly harder. 
Specifically, we wanted a form that is suited for scores that can be interpreted as confidence probabilities (that lie between 0 and 1), and capture the bias that $s_{ij}$ is high when $\tau_{ij}$ and $y$ agree, and low otherwise.  
For confidence variables, a natural parametric form of density is the beta density.  The beta density is popularly expressed in terms of two independent parameters $\alpha > 0$ and $\beta > 0$ as $P(s|\alpha,\beta) \propto s^{\alpha-1}(1-s)^{\beta-1}$.  Instead of independently learning these parameters, we chose an alternative representation that allows expression of user prior on the expected $s$. We write the beta in terms of two alternative parameters: the mean parameter $q_j^c$ and  the scale parameter $\pi_1$.  These are related to $\alpha$ and $\beta$ as $\alpha = q_j^c\pi_1$ and $\beta = (1-q_j^c)\pi_1$. 
 
We define our continuous potential as:
\begin{equation}
\label{eqn:contPots}
\psi_{\pi}(\tau_{ij}, s_{ij}, y) = 
\begin{cases}
    \textit{Beta}(s_{ij};\alpha_a, \beta_a )  & \text{if $k_j=y ~\& ~\tau_{ij} \ne 0$,} \\
    \textit{Beta}(s_{ij};\alpha_d, \beta_d )  & \text{if $k_j \ne y ~\& ~\tau_{ij} \ne 0$,} \\
    1 & \text{otherwise}
\end{cases}
\end{equation}
where $\alpha_a=q_j^c \pi_{jy}$ and $\beta_a=(1-q_j^c)\pi_{jy}$ are parameters of the agreement distribution and $\alpha_d=(1-q_j^c)\pi_{jy}$ and   $\beta_d=q_j^c \pi_{jy}$ are parameters of the disagreement distribution, where $\pi_{jy}$ is constrained to be strictly positive. To impose $\pi_{jy}>0$ while also maintaining differentiability, we reparametrize $\pi_{jy}$ as $\exp(\rho_{jy})$.  Thus, we require $K$ parameters for each continuous LF, which is the same as for a discrete LF. The Beta distribution would normally require $2K$ parameters but we used the user provided qualify guide in that special manner shown above to share  the mean  between the agreeing and disagreeing Beta. 

%
%
We experimented with a large variety of other potential forms before converging on the above.  We will elaborate on alternatives in the experimental section.  

With these potentials, the normalizer $Z_\theta$ of our joint distribution (Eqn~\ref{eq:joint}) can be calculated as 
\begin{equation}
\begin{split}
Z_\theta = & \sum_y \prod_j\sum_{\tau_{j} \in \{k_j, 0\}} 
 \psi_\theta(\tau_{j}, y)\int_{s_j=0}^1 \psi_\pi(\tau_{j}, s_j, y)
\\
         = & \sum_{y\in\cY}\prod_j(1+\exp(\theta_{jy}))
         \end{split}{}
\end{equation}
The normalizer reveals two further facets of our joint distribution. First our continuous potentials are defined such that when summed over $s_j$-s we get a value of 1, hence the normalizer is independent of the continuous parameters $\pi$.  That is, the continuous potentials $\psi_\pi(\tau_{ij}, s_{ij}, y)$ are locally normalized Bayesian probabilities $P(s_{ij}|\tau_{ij},y)$.  Second, the discrete potentials are not locally normalized; the  $\psi_\theta(\tau_{j}, y)$ cannot be interpreted as $\Pr(\tau_j|y)$ because by normalizing them globally we were able to learn the interaction among the LFs better. We will show empirically that either the full Bayesian model with potentials $P(y), P(\tau_{ij}|y)$, and $P(s_{ij}|\tau_{ij},y)$ or the fully undirected model where the $ \psi_\pi(\tau_{ij}, s_{ij}, y)$ potential is un-normalized are both harder to train.

\subsection{Training \model}
\label{sec:train}
Our training objective can be expressed as:
\begin{equation}
    \max_{\theta,\pi} LL(\theta,\pi|D) + R(\theta,\pi|\{q_j^t\})
    \label{eq:regularizedObjective}
\end{equation}
The first part maximizes the likelihood on the observed $\vtau_i$ and $\vs_i$ values of the training sample $D = \vx_1,\ldots,\vx_m$ after marginalizing out the true $y$. It can be expressed as:
\begin{align}
     & LL(\theta,\pi|D) = \sum\limits_{i=1}^{m} \log \sum\limits_{y \in \cY} P_{\theta,\pi}(\vtau_i,\vs_i, y) \label{eq:nonregulariedObjective} \\  
     & = \sum\limits_{i=1}^{m} \log \sum\limits_{y \in \cY}  \prod\limits_{j=1}^{n} \psi_j(\tau_{ij},y) \left(\psi_j(s_{ij}, \tau_{ij},y )\right)^{\text{cont}(\lambda_j)} - m \log Z_\theta \nonumber
\end{align}
By \modelnoconst, we will hereafter refer to the model in Eqn~\ref{eq:joint} that has parameters learnt by maximizing only this (first) likelihood part of the objective in Eqn~\ref{eq:nonregulariedObjective} and not the second part $R(\theta,\pi|\{q_j^t\})$. $R(\theta,\pi|\{q_j^t\})$ is a regularizer that guides the parameters with the programmer's expectation of the quality of each LF.  We start by motivating the need for the regularizer by showing simple cases that can cause the likelihood-only training to yield poor accuracy. 
\paragraph{Example 1: Sensitivity to Initialization} 
Consider a binary classification task where the $n$ LFs are perfect oracles  that trigger only on instances whose true label matches $k_j$.  Assume all $\lambda_j$ are discrete.  The likelihood of such data can be expressed as:
\begin{align}
    & LL(\theta) = \sum_{i=1}^m \log(\exp(\sum_{j:k_j=y} \theta_{j1}) + \exp(\sum_{j:k_j=y} \theta_{j2})) \nonumber \\ & - m\log(\prod_j(1+\exp(\theta_{j1})) + \prod_j(1+\exp(\theta_{j2}))) 
    \label{eq:oracle}
\end{align}
The value of the above likelihood is totally symmetric in  $\theta_{j1}$ and
$\theta_{j2}$ but the accuracy is not.  We will get 100\% accuracy only when the parameter for the agreeing case: $\theta_{jk_j}$ is larger than $\theta_{jy}$ for
$y \ne k_j$, and 0\% accuracy if $\theta_{jk_j}$ is smaller.  A trick
is to initialize the $\theta$ parameters carefully so that the
agreeing parameters $\theta_{jk_j}$ do have larger values.  However,
even such careful initialization can be forgotten in less trivial cases as we show in the next example.

%

\paragraph{Example 2: Failure in spite of good initialization} Consider a set S1 of $r$ LFs that assign a label of 1 and remaining set S2 of $n-r$ LFs that assign label 2. Let each true class-2 instance trigger one or more LF from S1 and one or more LF from S2.  Let each true class-1 instance trigger {\em only} LFs from S1.   When we initialize LFs in set S1 such that $\theta_{j1}-\theta_{j2}  > 0$ and LFs in set S2 have  $\theta_{j2}-\theta_{j1}  > 0$,  we can get good accuracy. However,
as training progresses the likelihood will be globally maximized when both sets of LFs favor the same class on all instances.  If we further assume that the true class distribution is skewed, the $LL(\theta)$ objective quickly converges to this useless maxima.  This scenario is not artificial. Many of the real datasets (e.g. the LFs of Spouse relation extraction data in Table~\ref{tab:sample}) exhibit such trends.


A straight-forward fix of the above problem is to impose a penalty on the sign of $\theta_{jk_j} - \theta_{jy}$. However, since the $\theta$s of LFs interact via the global normalizer $Z_\theta$ this condition is neither necessary nor sufficient to ensure that in the joint model $P_\theta(y,\vtau)$ the values of $y$ and $k_j$ agree more than disagree. For globally conditioned models the parameters cannot be easily interpreted, and we need to constrain at the level of the joint distribution.

One method to work with the whole distribution is to constrain the conditional $P_\theta(y|\vtau_i)$ over the instances where the LF triggers and constrain that the accumulated probability of the agreeing $y$ is at least $q_j^t$ as follows:
\begin{equation}
\label{dataDep:prec}
R(\theta|\{q_j^t\},D) = \sum_j \text{softplus}\left(\sum_{i:\tau_{ij}=k_j} (q_j^t - P_{\theta}(\tau_i,k_j))\right)   
\end{equation}
We call this the data-driven constrained training method and refer to it as \modeldep.
However, a limitation of this constraint is that
in a mini-batch training environment it is difficult to get enough examples per batch for reliable estimation of the empirical accuracy, particularly for LFs that trigger infrequently.
Next we present our method of incorporating the user guidance into the trained model to avoid such instability.


\subsection{Data-independent quality guides in \model}
\label{sec:const}
Our final approach that worked reliably was to regularize the parameters so that the learned joint distribution of $y$ and $\tau_j$ matches the user-provided quality guides $q_j^t$  over all $y,\tau_{j}$ values from the joint distribution $P_{\theta,\pi}$. By default, this is the regularizer that we employ in \mbox{CAGE}. 

The $q_j^t$ guide is the user's belief on the fraction of cases where $y$ and $\tau_{j}$ agree when $\tau_{j} \ne 0$  (LF $\lambda_j$ triggers).  Using the joint distribution we can calculate this agreement probability as $P_\theta(y = k_j |\tau_j = k_j)$.
This probability can be computed in closed form by marginalizing over all remaining variables in the model in Equation~\ref{eq:joint} as follows:
\newcommand{\precj}{P_\theta(y=k_j|\tau_j = k_j)}
\begin{eqnarray*}
    \precj &=& \frac{P_\theta(y=k_j, \tau_j = k_j)}{P_\theta(\tau_j = k_j)} \\
    &=&  \frac{\msg_j(k_j)\prod_{r\ne j} (1+\msg_r(k_j))}
    {\sum_{y \in \cY} \msg_j(y)\prod_{r\ne j} (1+\msg_r(y))}
\end{eqnarray*}
where $\msg_j(y)=\exp(\theta_{jy})$.  We then seek to minimize the KL distance between the user provided $q_j^t$ and the model calculated precision $\precj$ which turns out to be:  
\begin{equation}
\begin{split}
        R(\theta | \{q_j^t\}) = \sum_j  q_j^t \log \precj \\ 
        + (1-q_j^t)  \log (1-\precj)
\end{split}
\label{eq:userSummLLL}
\end{equation}
Specifically, when the \model model is restricted only to discrete LFs while also incorporating the quality guide in Eqn~\ref{eq:userSummLLL} into the objective in Eqn~\ref{eq:nonregulariedObjective}, we refer to the approach as \modelnocont. Further, when the quality guide in Eqn~\ref{eq:userSummLLL} is dropped from \modelnocont, we refer to the approach as \modelnoconstnocont.

\subsection{Relationship of \model with existing models}
\label{sec:compare}
We would like to point out that the following two simplifications in \model lead to existing well known models~\citep{snorkel_nips2016,snorkel_vldb2018}, {\em viz.}, 
(i) Coupling the $\theta_{yj}$ parameters,
(ii) Ignoring quality guides, and (iii) Not including continuous potentials.  
The design used in \cite{snorkel_icml2017} is to assign a single parameter $\theta_j$ for each LF and share it across $y$ as:
\begin{equation}
\psi_{j}^\text{snorkel}(\tau_{ij},y) = 
\begin{cases}
    \exp(\theta_{j})  & \text{if $\tau_{ij}\ne 0, y = k_j$,} \\
     \exp(-\theta_{j})  & \text{if $\tau_{ij}\ne 0, y \ne k_j$,} \\
    1 & \text{otherwise.}
\end{cases}
\label{eq:snorkeloriginal}
\end{equation}
After ignoring quality guides and continuous LF, we note that a choice of $\theta_{j,+1} = - \theta_{j,-1}$ makes \model exactly same as the model in Snorkel.
However, we found the Snorkel's method of parameter sharing incorporates an unnecessary bias that $P_\theta(\tau_{ij} = 0|y=k_j) = 1 - P_\theta(\tau_{ij} = 0|y \ne k_j)$. 
Also, Snorkel's pure likelihood-based training is subject to all the sensitivity to parameter initialization and training epochs that we highlighted in Section~\ref{sec:train}.  We show in the experiments how each of the three new extensions in \model is crucial to getting reliable training with the data programming paradigm.


\section{Empirical Evaluation}
\label{exp}
In this section we (1) evaluate the utility of continuous LFs vis-a-vis discrete LFs, (2) demonstrate the role of the quality guides for the stability of the unsupervised likelihood training of \model as well as Snorkel, and (3) perform a detailed ablation study to justify various design elements of our generative model and its guided training procedure.
\subsection{Datasets and Experiment Setup}
\label{sec:setup}
We perform these comparisons on five different datasets. 

\noindent
{\bf Spouse:}~\citep{spouse}
This is a relation extraction dataset that proposes to label candidate pairs of entities in a sentence as expressing a `spouse' relation or not. 
Our train-dev-test splits and set of discrete LFs shown in Table~\ref{tab:sample} are the same as in~\citep{snorkel_nips2016} where it was first used.  
For each discrete LF that checks for matches in a dictionary $D$ of keywords we create a continuous LF that returns $s_j$ as the maximum of cosine similarity of their word embeddings as shown in Table~\ref{tab:continuoussample}.
We used pre-trained vectors provided by Glove~\cite{glove}.

\noindent
{\bf SMS spam}~\citep{smsspam}  is a binary spam/no-spam classification dataset with 5574 documents split into 3700 unlabeled-train and
1872 labeled-test instances. 
Nine LFs are created based on (i) presence of three categories of words which are highly likely to indicate spam (ii) presence of 2 categories of trigger words in certain contexts, (iii) reference to keywords indicative of first/second or third person, (iv) text characteristics such as number of capitalized characters, presence of special characters, {\em etc.} and finally a LF that is (v) associated with the negative class, always triggers and serves as the class prior. The LFs are explained in the supplementary in Section~\ref{sec:smsspamlfs}. 
The  continuous LFs are created in the same way as in Spouse based on word-embedding similarity, number of capitalized characters, {\em etc.}

\noindent
{\bf CDR:}~\citep{cdr} This is also a relation extraction dataset where the task is to detect whether or not a sentence expresses a `chemical cures disease' relation. 
The train-dev-test splits and LFs are the same as in \citep{snorkel_nips2016}. We did not develop any continuous LF for CDR.

\noindent
{\bf Dedup:} This dataset{Publicly available at \url{https://www.cse.iitb.ac.in/~sunita/alias/}} comprises of 32 thousand pairs of noisy citation records with fields like Title, Author, Year {\em etc.} The task is to detect if the record pairs are duplicates. We have 18 continuous LFs corresponding to various text similarity functions (such as Jaccard, TF-IDF similarity, 1-EditDistance, {\em etc.}) computed over one or more of these fields. Each of these LFs is positively correlated with the duplicate label; we create another 18 with the score as 1-similarity for the negative class. The dataset is highly skewed with only 0.5\% of the instances as duplicate. All LFs here are continuous.

\noindent
{\bf Iris:}
Iris is a UCI dataset with 3 classes. We split it into 105 unlabeled train and 45 labeled test examples.  We create LFs from the 4 features of the data as follows: For each feature $f$ and class $y$, we calculate $f$'s mean value $\overline{f}_y$ amongst the examples of $y$ from labeled data and create a LF that returns the value $1-\text{norm}(f-\overline{f}_y)$  - where $\text{norm}(f-\overline{f}_y)$ is the normalized distance from this mean. This gives us a total of $4\times 3 = 12$ continuous LFs. Each such LF has a corresponding discrete LF that is triggered if the feature is closest to the mean of its corresponding class. 

\noindent
{\bf Ionosphere:}
This is another 2-class UCI dataset, that 
is split to 245 unlabeled train and 106 labeled test instances. 64 continuous LFs are created in a manner similar to Iris.

\noindent
{\bf Training Setup} For each dataset and discrete LF we arbitrarily assigned a default discrete quality guide $q_j^t=0.9$ and for continuous LFs $q_j^c=0.85$. We used learning rate of $0.01$ and $100$ training epochs. Parameters were initialized  favorably --- so for the agreeing parameter initial $\theta_{jk_j=1}$ and for disagreeing parameter initial $\theta_{ky}=-1, y \ne k_j$. For Snorkel this is equivalent to $\theta_j=1$. Only for \model\ that is trained with guides we initialize all parameters to 1.  We show in Section~\ref{sec:init} that \model is insensitive to initialization whereas others are not.

\noindent
{\bf Evaluation Metric}
We report F1 as our accuracy measure on all binary datasets and for the multi-class dataset Iris we measure micro F1 across the classes.   From our generative model, as well as Snorkel, the predicted label on a test instance $\vx_i$ is the $y$ for which the joint generative probability is highest, that is: $\text{argmax}_y P(y,\vtau_{i},\vs_{i})$.   Another measure of interest is the accuracy that would be obtained by training a standard discriminative classifier $P_\vw(y|\vx)$ with labeled data as the probabilistically labeled $P(y|\vx_i) \propto P(y,\vtau_{i},\vs_{i})$ examples $\vx_i$ from the generative model.  In the first part of the experiment we measure the accuracy of labeled data produced by the generative model. In Section~\ref{sec:disc} in the Supplementary,  we present accuracy from a trained discriminative model from such dataset. 
We implemented our model in Pytorch.\footnote{Code available at \url{https://github.com/oishik75/CAGE}.}



\subsection{Overall Results}
\begin{table}[!htb]
\setlength\tabcolsep{4.0pt}
{\begin{tabular}{|l|r|r|r|r|r|r|}
\hline &  Spouse &  CDR    &  SMS    &  Ion &  Iris  &  Dedup  \\ \hline
 Majority  & 0.17  & 0.53   &  0.23    &  0.79       & 0.84  & -  \\ \hline
 Snorkel  & 0.41 & 0.66   & 0.34   & 0.70       & 0.87 & -      \\ \hline
 \modelnoconstnocont & 0.48 & 0.69 & 0.34   & 0.81     & 0.87 & -      \\ \hline
 \modelnocont & 0.50 & 0.69 & 0.45 & 0.82      & 0.87 & -      \\ \hline
 \model & {\bf 0.58} & {\bf 0.69}  & {\bf 0.54}  & {\bf 0.97}  &   {\bf 0.87} & {\bf 0.79} \\ \hline
\end{tabular}}
\caption{Overall Results (F1) with predictions from various generative models contrasted with the Majority baseline.}
\label{tab:overall}
\end{table}

In Table~\ref{tab:overall}, we compare the performance of \model in terms of F1, against the following alternatives: (i) {\bf Majority:} This is a simple baseline, wherein, the label on which a majority of the LFs show agreement is the inferred consensus label. 
(ii) {\bf Snorkel:} See Section~\ref{sec:compare}. 
(iii) {\bf \modelnoconstnocont:} Our model without continuous LFs and quality guides (See Section~\ref{sec:const}).  
(iv) {\bf \modelnocont:} Our model with quality guides but without continuous LFs (See Section~\ref{sec:const}).  

From this table we make three important observations: (1)
Comparing Snorkel with \modelnoconstnocont\ that differs only in decoupling Snorkel's shared $\theta_{j}$ parameters, we observe that the shared parameters of Snorkel were indeed introducing undesirable bias.
(2) Comparing \modelnoconstnocont\ and \modelnocont\ we see the gains due to our quality  guides.  (3) 
Finally, comparing \modelnocont\ and \model we see the gains because of  the greater expressibility of continuous LFs. These LFs required negligible additional human programming effort beyond the discrete LFs.  Compared to Snorkel our model provides significant overall gains in F1.  For datasets like Dedup which consist only of continuous scores, \model\ is the only option.

We next present a more detailed ablation study to tease out the importance of different design elements of \modelNS.  

\subsection{Role of the Quality Guides}
\label{sec:init}
\begin{figure*}[!ht]
\centering
\begin{tabular}{ccc}
\includegraphics[width=0.33\textwidth]{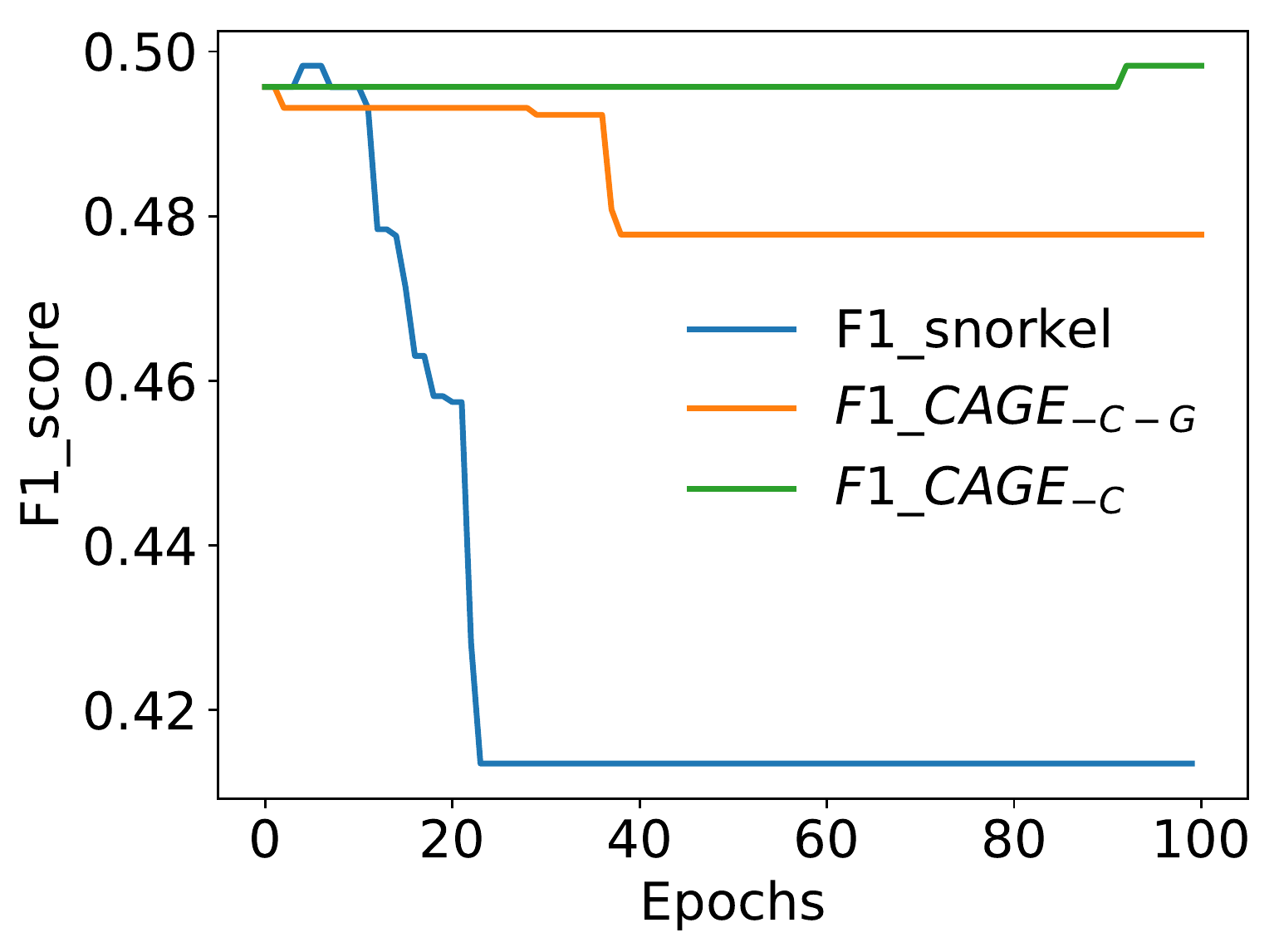} &  
\includegraphics[width=0.33\textwidth]{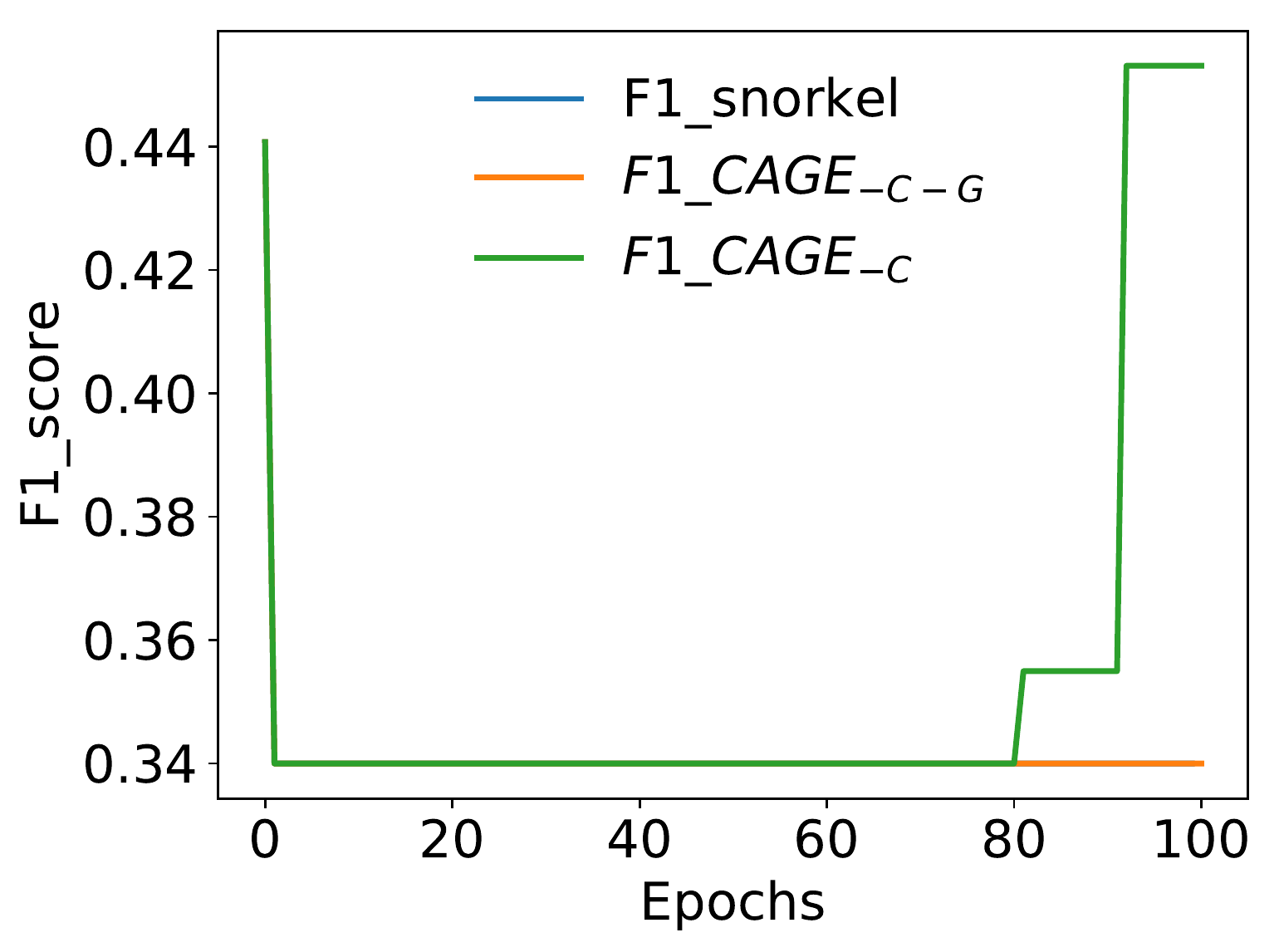} &
\includegraphics[width=0.33\textwidth]{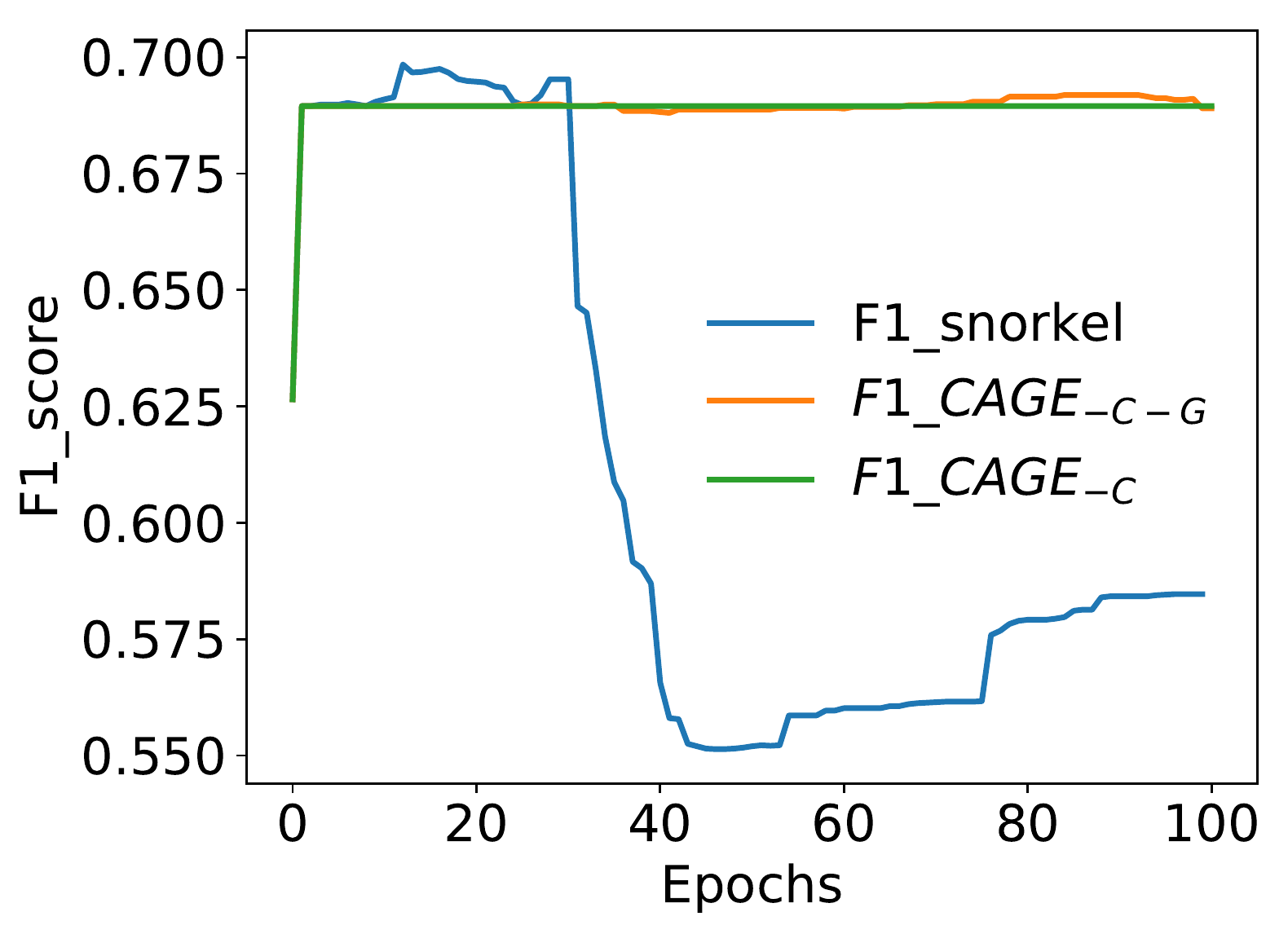}
\\
(a)~Spouse  & (b) SMS  & (c) CDR 
\end{tabular}
\caption{F1 with increasing number of training epochs compared across snorkel, \modelnoconstnocont and \modelnocont, for three datasets. For each dataset, in the absence of guides, we observe unpredictable variation in test F1 as training progresses.}
\label{fig:prec}
\end{figure*}
We motivate the role of the quality guides in Figure~\ref{fig:prec} where we show test F1 for increasing training epochs on three datasets. In these plots we considered only discrete LFs.
We compare Snorkel and our model with (\modelnocont) and without (\modelnoconstnocont) these guides.   
Without the quality guides, all datasets exhibit unpredictable swings in test F1.  These swings cannot be attributed to over-fitting since in Spouse and SMS F1 improves later in training with our quality guides.  Since we do not have  labeled validation data to choose the correct number of epochs, the quality guides are invaluable in getting reliable accuracy in unsupervised learning.

\begin{figure}[ht]
\centering
\includegraphics[width=0.35\textwidth]{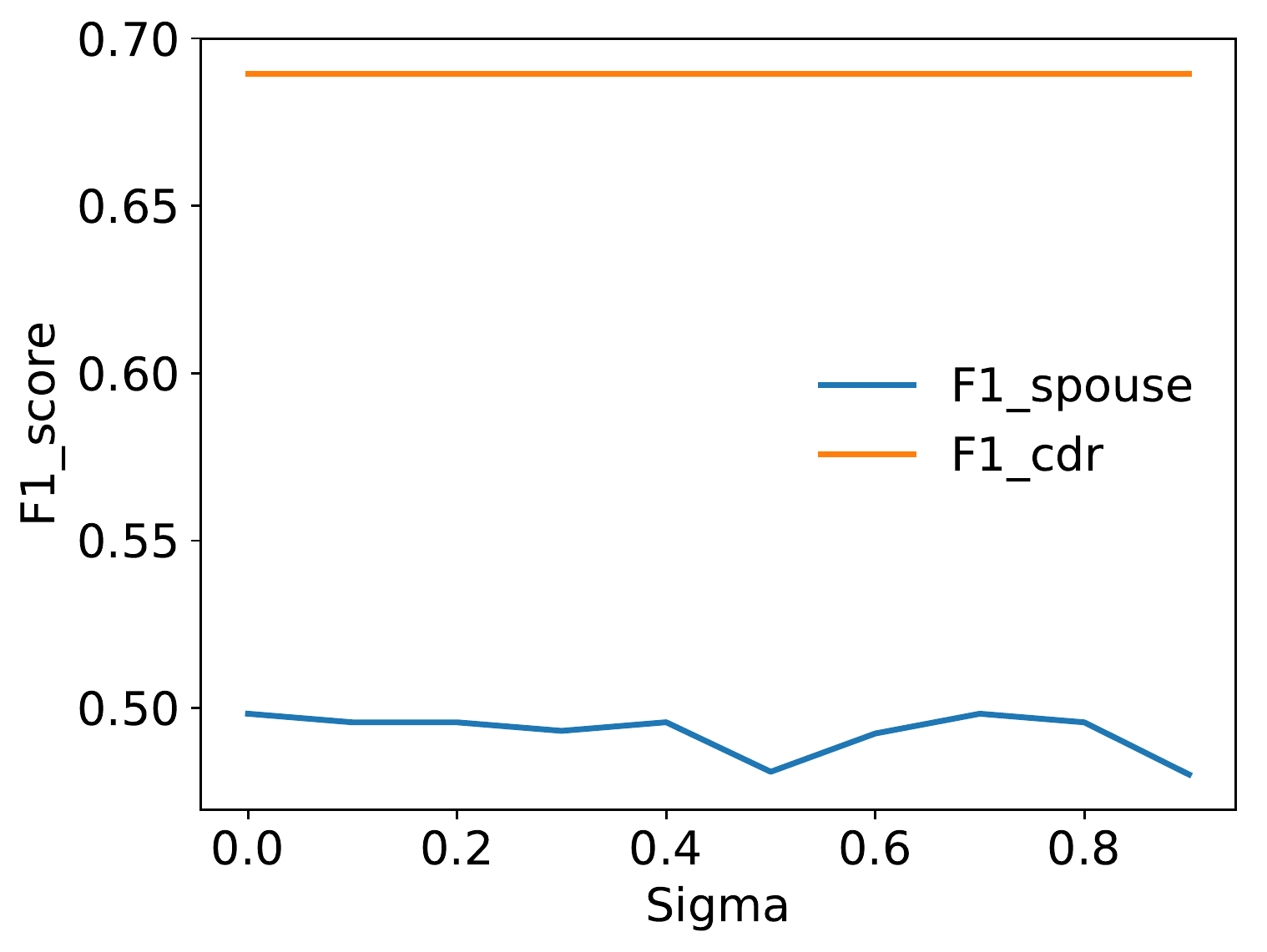}
\caption{F1 with increasing distortion in the guess of the LF quality guide,  $q^t_j$.}
\label{fig:distort}
\end{figure}

Next, we show that the stability provided by the quality guides ($q^t_j$) is robust to large deviations from the true accuracy of a LF.  Our default $q^t_j$ value was 0.9 for all LFs irrespective of their true accuracy.  We 
repeated our experiments with a precision of 0.8 and got the same accuracy across training epochs (Figure~\ref{fig:stable} in  Supplementary).  We next ask if knowing the true accuracy of a LF would help even more and how robust our training is  to distortion in the user's guess from the true accuracy.  We calculated true accuracy of each LF on the devset and distorted this by a Gaussian noise with variance $\sigma$. In Figure~\ref{fig:distort} we present accuracy after 100 epochs on two datasets with increasing distortion ($\sigma$).  On CDR \modelNS's accuracy is very robust to distorted $q^t_j$ but guides are important as we see from Figure~\ref{fig:prec}(c).   On Spouse accuracy is highest with perfect values of $q^t_j$ (Sigma=0) but it stays close to this accuracy up to a distortion of 0.4.

\noindent
{\bf Sensitivity to Initialization:} We carefully initialized parameters of all models except \model.  With random initialization all models without guides (Snorkel and \modelnoconst) provide very poor accuracy.  Exact numbers are in the supplementary material ({\it c.f.} Figure~\ref{fig:init}).

\begin{table}[!htb]
\resizebox{\columnwidth}{!}
{\begin{tabular}{|l|l|l|l|l|}
\hline & \bf Spouse & \bf CDR    & \bf Sms    & \bf Ion \\ \hline
\modelpenalty & 0.48  & 0.69  & 0.34   & 0.81      \\ \hline
\modelnoconstnocont & 0.48  & 0.69  & 0.34   & 0.81      \\ \hline
\modeldepnocont & 0.48  & 0.69  & 0.34   & 0.81      \\ \hline
\modelnocont & {\bf 0.50} & {\bf 0.69} & {\bf 0.45} & {\bf 0.82}      \\ \hline
\end{tabular}}
\caption{Comparing different methods of incorporating user's quality guide on discrete LFs.}
\label{tab:role_of_constraints}
\end{table}
\begin{table*}[]
\end{table*}

\paragraph{Method of Enforcing Quality Guides}
In Table~\ref{tab:role_of_constraints}, we compare F1 
for the following choices:
    (i) {\bf \modelpenalty}: Our model with the objective in Eqn (\ref{eq:nonregulariedObjective}) augmented with the sign penalty $\max(0,\theta_{jy \ne k_j} - \theta_{jk_j})$  instead of the regularizer in Eqn (\ref{eq:userSummLLL}). 
    (ii) {\bf \modelnoconstnocont}: Our model without guides
    (iii) {\bf \modeldepnocont}: The data-driven method of incorporating quality guides (See Section~\ref{sec:train}), 
    and (iv) {\bf \modelnocont}: our data independent regularizer of the model's marginals with $q_j^t$ (Eqn~\ref{eq:userSummLLL}). 
%
From  Table~\ref{tab:role_of_constraints} we observe that  \modelnocont\ is the only one that provides reliable gains. Thus, it is not just enough to get quality guides from users, we need to also design sound methods of combining them in likelihood training. 

\subsection{Structure of the Potentials}
In defining the joint distribution $P_{\theta,\pi}(y,\vtau_{i},\vs_i)$
(Eq~\ref{eq:joint}) we used undirected globally normalized potentials
for the discrete LFs(Eqn~\ref{eq:decoupledthetas}).  We compare with
an alternative where our joint is a pure directed Bayesian network
with potentials $Pr(\tau_j|y)=\exp(\theta_{jy}/(1+\exp(\theta_{jy})$
on each discrete LF and a class prior $\Pr(y)$.  We observe that the
undirected model is better able to capture interaction among the LFs
with the global normalization $Z_\theta$.
\begin{tabular}{|l|l|l|l|l|l|}
\hline
         & \textbf{Spouse} & \textbf{CDR} & \textbf{Sms} & \textbf{Ion} & \textbf{Iris} \\ \hline
Directed & 0.15            & 0.49            & 0.59         & 0.86                & {\bf 0.89}          \\ \hline
\model     & {\bf 0.58}            & {\bf 0.69}         & {\bf 0.54}         & {\bf 0.97}                & 0.87          \\ \hline
\end{tabular}

\noindent
We repeated other  ablation experiments where  the continuous potentials are undirected and take various forms.  The results appear in the Supplementary in Sect~\ref{sec:contPotsAblation}
and show that local normalization is crucially important for modeling $s_j$ of continuous LFs.

\subsection*{Related Work}
\label{relwork}

Several consensus-based prediction combination algorithms~\citep{gao2009graph,kulkarniaaai2018}
exist that combine multiple model predictions to counteract the effects of data quality and model bias. 
There also exist label embedding approaches from the extreme classification literature~\citep{yeh2017learning}
that exploit inter-label correlation. 
While these approaches assume that the imperfect labeler's knowledge is fixed, \cite{fang2012self} present a self-taught active learning paradigm, where a crowd of imperfect labelers learn complementary knowledge from each other. However, they use instance-wise reliability of labelers to query only the most reliable labeler without any notion of consensus. A recent work by~\cite{chang2017revolt} presents a collaborative crowd sourcing approach. However, they are motivated by the problem of eliminating the burden of defining labeling guidelines a priori and their approach harnesses the labeling disagreements to identify ambiguous concepts and create semantically rich structures for post-hoc label decisions. 

There is work in the crowd-labeling literature that makes use of many imperfect labelers~\citep{kulkarniaaai2018,raykar2010learning,yan2011active,Dekel:2009:GLE:1553374.1553404} and accounts for both labeler and model uncertainty to propose probabilistic solutions to (a) adapt conventional supervised learning algorithms to learn from multiple subjective labels; (b) evaluate them in the absence of absolute gold standard; (c) estimate reliability of labelers. \cite{Donmez:2008:PLC:1458082.1458165} propose a \emph{proactive learning} method that jointly selects the optimal labeler and instance with a decision theoretic approach. Some recent literature has also studied the augmenting neural networks with rules in first order logic to either guide the individual layers~\cite{acl2019augmentinglogic} or  train model weights within constraints of the rule based system using a student and teacher model~\cite{Hu2016HarnessingDN}

Snorkel~\citep{snorkel_nips2016,snorkel_icml2017,snorkel_vldb2018,snorkel_acl2018,snorkel_icml2019} relies
on domain experts manually developing heuristic and noisy LFs. Similar methods that rely on imperfect sources of labels are~\cite{Bunescu:Mooney:07,Hearst:1992:AAH:992133.992154} relying on heuristics, ~\cite{Mintz:2009:DSR:1690219.1690287} on distant supervision and ~\cite{ruleensembles} on learning conjunctions discrete of (discrete) rules. 
The aforementioned literature focuses exclusively on labeling suggestions that are discrete.
We present a generalized generative model to aggregate heuristic labels from continuous (and discrete) LFs while also incorporating user accuracy priors. 



\paragraph{Conclusion}
\label{conc}
We presented a data programming paradigm that lets the user specify labeling functions  which when triggered on instances can also produce continuous scores.  The unsupervised task of  consolidating weak labels is inherently unstable and sensitive to parameter initialization and training epochs.  Instead of depending on un-interpretable hyper-parameters which can only be tuned with labeled validation data which we assume is unavailable, we let the user guide the training with interpretable quality guesses. We carefully designed the potentials and the training process to give the user more interpretable control. 

\paragraph{\bf Acknowledgements} This research was partly sponsored by a Google India AI/ML Research Award and partly sponsored by IBM Research, India (specifically the IBM AI Horizon Networks - IIT Bombay initiative).
\bibliographystyle{aaai}
\bibliography{aaai2020}
\newpage
\onecolumn
{\huge{\bf Data Programming using Continuous and Quality-Guided Labeling Functions (Supplementary Material)}}

\hspace{2cm}

\section{Challenges in modeling continuous LFs: Illustration through an Example}
\label{sec:motivationExample}

\begin{table}[h!]
\centering
\begin{tabular}{| c | c | c | c | c | } 
 \hline
 Sentence id  & \multicolumn{2}{|c|}{Discrete} & \multicolumn{2}{|c|}{Continuous} \\ [0.5ex] 
    & LF1 (+1) & LF2 (-1) & LF1 (+1) & LF2 (-1) \\ [0.5ex] 
 \hline
$S_1$ (+1) & 1 & 0 & \(1, ~1.0\) & \(1, ~0.84\)  \\ 
$S_2$ (-1) & 0 & 1 & \(1, ~0.75\) & \(1, ~1.0\)  \\ 
 \hline
\end{tabular}
\caption{Triggering of the continuous and discrete versions of functions LF1 and LF2 for the two example sentences $S_1$ and $S_2$.}
\label{table:twochallenges}
\end{table}
In this section, we highlight some challenges that we attempt to address while modeling continuous LFs. We illustrate the challenges through two example sentences. Beside each sentence id, we state the value of the true label ($\pm 1$). While the candidate pair in $S_1$ is an instance of the  spouse relationship, the pair in $S_2$ is not: \\
$\langle S_1,+1 \rangle$: It's no secret that Brandi Glanville's relationship with ex-husband $<$Eddie Cibrian$>$ and his wife, $<$LeAnn Rimes$>$, has been a strained one -- but Glanville tells ETonline it's getting better. \\
$\langle S_2,-1\rangle$: Afterwards, $<$Christian Brady$>$, Dean of the Schreyer Honors College and father of $<$Mack Brady$>$, whom the game is named after, addressed the lacrosse team.  \\\

In Table~\ref{table:twochallenges}, we present the outputs of both discrete and continuous LFs on the examples $S_1$ and $S_2$. For $S_1$,  in the discrete case, correct consensus can easily be performed to output the true label +1 as LF1 (designed for class +1) gets triggered whereas LF2 (designed for class -1) is not triggered. 
correct consensus is challenging since both LF1 and LF2 produce non-trivial scores. Let us say that we threhold the score of each continuous LF to be able to mimic the discrete case; {\em e.g.}, score above the threshold will mean that the LF is triggered and otherwise, not.  However, the threshold would depend on the particular LF (possibly 0.75 for LF1 and 0.84 for LF2), and is tricky to estimate in an unsupervised setup such as ours. To address this challenge, we significantly extend and modify the existing generative model~\citep{snorkel_nips2016,snorkel_icml2017} to support continuous LFs. In addition to modeling the triggering distribution of each LF (continuous or discrete), we also model the distribution of scores for each continuous LF.

\section{Robustness to Parameter Initialization}
\label{sec:init}
\begin{figure*}[!ht]
\centering
\begin{tabular}{cc}
\includegraphics[width=0.33\textwidth]{spouse_f1.pdf} &  
\includegraphics[width=0.33\textwidth]{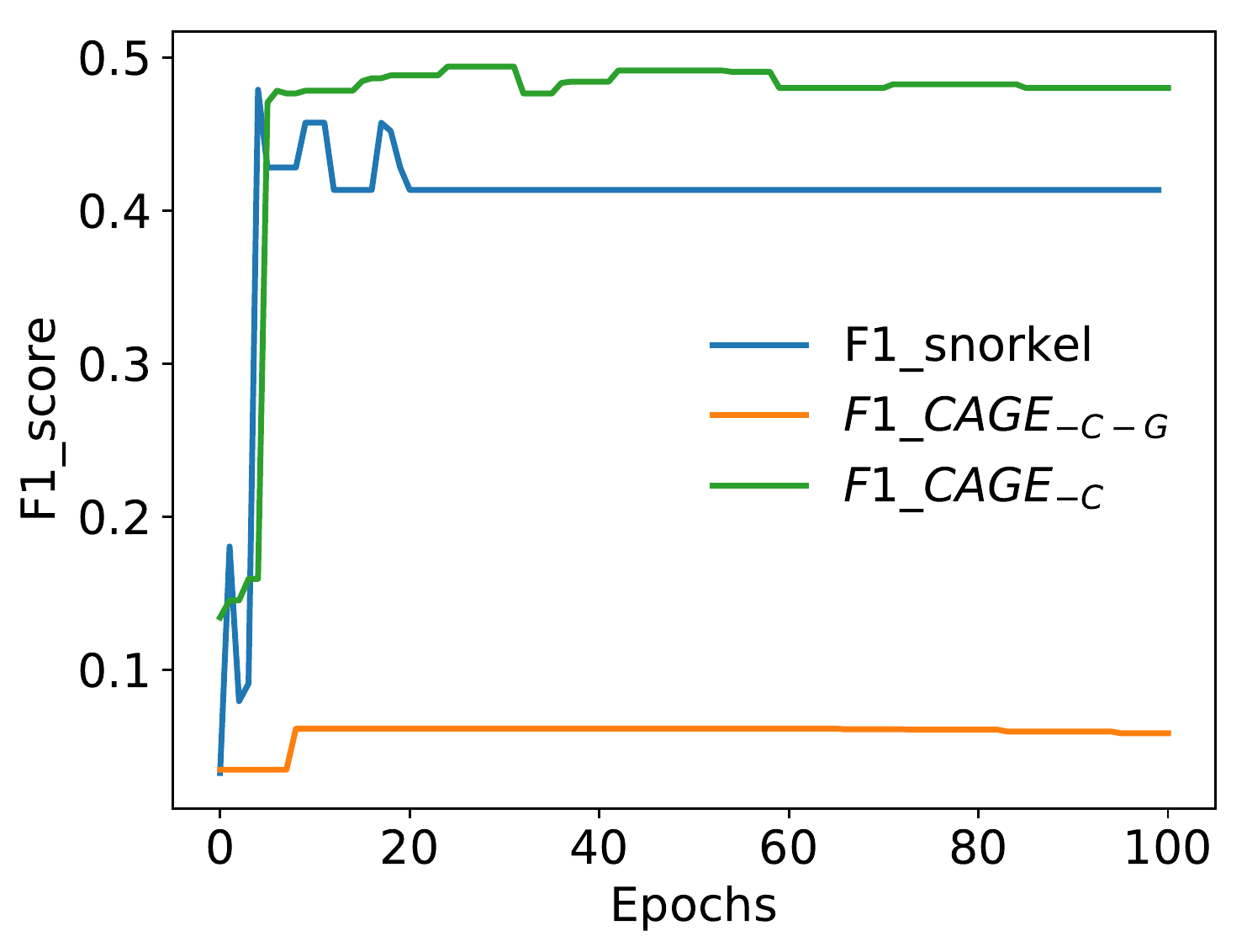} \\
\includegraphics[width=0.33\textwidth]{cdr_f1.pdf} &
\includegraphics[width=0.33\textwidth]{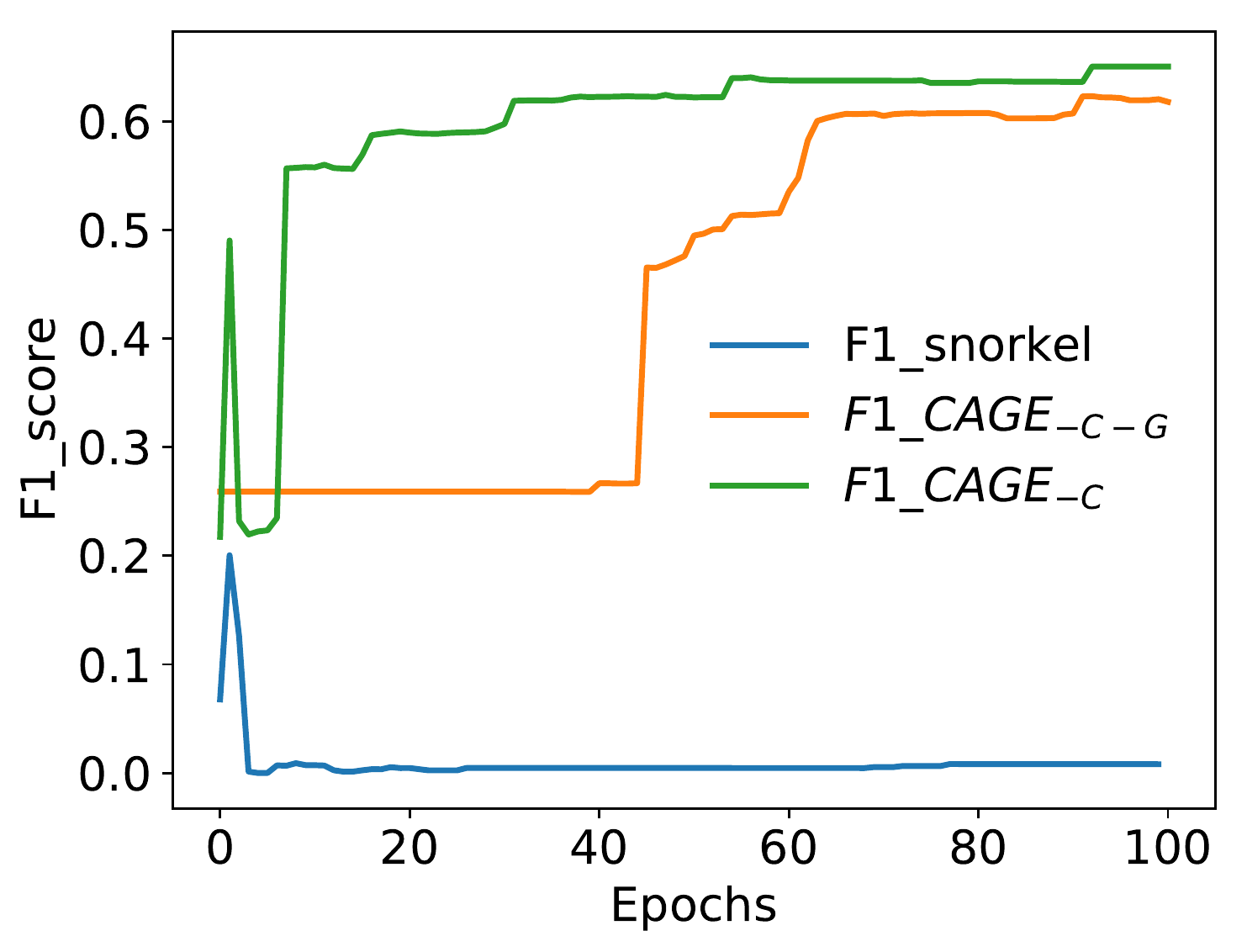} 
\\
(a)~Agreeing initialization  &  (b) Random Initialization. 
\end{tabular}
\caption{F1 with increasing number of training epochs compared across snorkel, \modelnoconstnocont and \modelnocont, for two datasets: Spouse (top-row) and CDR(bottom-row)}
\label{fig:init}
\end{figure*}

We trained our models with random Gaussian (0,0.1) initialization and compared with our agreeing initialization.
In Figure~\ref{fig:init} we show how \modelnocont is able to recover from any initialization whereas methods without guides fare even worse with random initialization.

\section{Labeling Functions employed in the SMS-Spam dataset}
\label{sec:smsspamlfs}

The SMS Spam dataset\footnote{\url{http://www.dt.fee.unicamp.br/~tiago/smsspamcollection/}} is a collection of raw SMS text. The task is to classify each SMS as spam or not. We employed 9 discrete and 6 continuous labeling functions. We briefly discuss each labeling function that we employed in this task, beginning with the discrete labeling functions:
\begin{enumerate}
    \item Three labeling functions based on the presence of trigger words which are highly likely to indicate spam. Each labeling function below, associated with the `spam' class is triggered if any of the words in the SMS matches at least one of the words associated with that LF
    \begin{enumerate}
	\item  LF1: matches one of the trigger words in the set trigWord1 = \{'free','credit','cheap','apply','buy','attention','shop','sex','soon',  'now','spam'\}. 
\begin{verbatim}
def LF1(c):
 return (1,1) if len(trigWord1.intersection(c['text'].split())) > 0 
 else (0,0)
\end{verbatim}
	\item  LF2: matches one of the trigger words in the set trigWord2 = \{'gift','click','new','online','discount','earn','miss','hesitate',  'exclusive','urgent'\}. 

\begin{verbatim}
def LF2(c):
 return (1,1) if len(trigWord2.intersection(c['text'].split())) > 0 
 else (0,0)
\end{verbatim}	
\item  LF3: matches one of the trigger words in the set trigWord3 = \{'cash','refund','insurance','money','guaranteed','save','win','teen', 'weight','hair'\}. 
%

\begin{verbatim}
def LF3(c):
 return (1,1) if len(trigWord3.intersection(c['text'].split())) > 0 
 else (0,0)
\end{verbatim}

   \end{enumerate}
    \item Two labeling functions associated with the `spam' class based on the presence of words/phrases in other parts of speech along with some {\bf context}, which are highly likely to indicate spam: 
    \begin{enumerate}
	\item  LF4: matches inquiries in the {\bf contex} of a person's being free or not free along with some overlap with the set of words notFreeWords = \{'toll','Toll','freely','call','meet','talk','feedback'\},  
\begin{verbatim}
def LF4(c):
 return (-1,1) if 'free' in c['text'].split() and 
  len(notFreeWords.intersection(c['text'].split()))>0 
 else (0,0)        
\end{verbatim}

	\item  LF5: matches inquiries in the {\bf contex} of a person's being free or not free along with substring match with the set of words/phrases notFreeSubstring = \{'not free','you are','when','wen'\},  
\begin{verbatim}
def LF4(c):
 return (-1,1) if 'free' in c['text'].split() and 
  re.search(ltp(notFreeSubstring),c['text'], flags= re.I) 
 else (0,0)                
\end{verbatim}

   \end{enumerate}

    \item Two labeling functions associated with the `spam' class based on the presence of first, second or third  person keyword matches: 
    \begin{enumerate}
	\item  LF6: matches first and second person keywords in firstAndSecondPersonWords = \{'I','i','u','you','ur','your','our','we','us','you\'re,'\},  
\begin{verbatim}
def LF6(c):
 return (-1,1) if 'free' in c['text'].split() and 
  len(person1.intersection(c['text'].split()))>0 
 else (0,0)        
\end{verbatim}

	\item  LF7: matches third person keywords in thirdPersonWords = \{'He','he','She','she','they','They','Them','them','their','Their'\},  
\begin{verbatim}
def LF7(c):
 return (-1,1) if 'free' in c['text'].split() and 
  len(thirdPersonWords.intersection(c['text'].split()))>0 
 else (0,0)        
\end{verbatim}

   \end{enumerate}

    \item Function based on text characteristics such as number of capitalized characters. The function below is triggered if the number of capital characters exceeds a threshold (6 in this case).
\begin{verbatim}
def LF8(c):
 return (1,1) 
  if (sum(1 for ch in c['text'] if ch.isupper()) > 6)
 else (0,0)
\end{verbatim}

    \item Class Prior labeling function: The function below is always triggered for the negative class and serves as a class prior.
    \begin{verbatim}
def LF9(c):
 return (-1,1)
\end{verbatim}

\end{enumerate}

All of our continuous labeling functions are continuous versions of the discrete  labeling functions described above. For those discrete labeling functions that match dictionary entries, the continuous counterpart computes the maximum word vector based similarity of the text (or textual context) across all entires in the dictionary. They are listed below:
\begin{enumerate}[label=(\roman*)]
    \item Three labeling functions based on the presence of trigger words which are highly likely to indicate spam:
    \begin{enumerate}
	\item  CLF1: Continuous version of LF1:
\begin{verbatim}
def CLF1(c):
 sc = 0
 word_vectors = get_word_vectors(c['text'].split())
 for w in trigWord1:
  sc=max(sc,get_similarity(word_vectors,w))
 return (1,sc)
\end{verbatim}
	\item  CLF2: Continuous version of LF2:
\begin{verbatim}
def CLF2(c):
 sc = 0
 word_vectors = get_word_vectors(c['text'].split())
 for w in trigWord2:
  sc=max(sc,get_similarity(word_vectors,w))
 return (1,sc)
\end{verbatim}
	\item  CLF3: Continuous version of LF3:
\begin{verbatim}
def CLF3(c):
 sc = 0
 word_vectors = get_word_vectors(c['text'].split())
 for w in trigWord3:
  sc=max(sc,get_similarity(word_vectors,w))
 return (1,sc)
\end{verbatim}

   \end{enumerate}
    \item Two labeling functions associated with the `spam' class based on the presence of words/phrases in other parts of speech along with some {\bf context}, which are highly likely to indicate spam: 
    \begin{enumerate}
	\item  CLF4: Continuous version of LF4:
\begin{verbatim}
def CLF4(c):
 sc = 0
 word_vectors = get_word_vectors(c['text'].split())
 for w in notFreeWords:
  sc=max(sc,get_similarity(word_vectors,w))
 return (1,sc)
\end{verbatim}

	\item  CLF5: Continuous version of LF5: 
\begin{verbatim}
def CLF5(c):
 sc = 0
 word_vectors = get_word_vectors(c['text'].split())
 for w in notFreeSubstring:
  sc=max(sc,get_similarity(word_vectors,w))
 return (1,sc)
\end{verbatim}

   \end{enumerate}

    \item Two labeling functions associated with the `spam' class based on the presence of first, second or third  person keyword matches: 
    \begin{enumerate}
	\item  CLF6: Continuous version of LF6:  
\begin{verbatim}
def CLF6(c):
 sc = 0
 word_vectors = get_word_vectors(c['text'].split())
 for w in firstAndSecondPersonWords:
  sc=max(sc,get_similarity(word_vectors,w))
 return (-1,sc)
\end{verbatim}

	\item  CLF7: Continuous version of LF7
\begin{verbatim}
def CLF7(c):
 sc = 0
 word_vectors = get_word_vectors(c['text'].split())
 for w in thirdPersonWords:
  sc=max(sc,get_similarity(word_vectors,w))
 return (-1,sc)
\end{verbatim}

   \end{enumerate}

    \item Continuous version of LF8; the value it returns increases with the number of capital characters:
\begin{verbatim}
def CLF8(c):
 l = sum(1 for ch in c['text'] if ch.isupper()) 
 return (1, 1-np.exp(float(-l/2)))
\end{verbatim}

    \item Class Prior labeling function: This remains the same as before.
\begin{verbatim}
def CLF8(c):
 return (-1,1)
\end{verbatim}

\end{enumerate}

\section{Form of continuous potential}
\label{sec:contPotsAblation}
We explored many different generic forms of potentials for continuous LFs and we compare them with the specific beta-distribution form.  These potentials $\psi(s_j, \tau_j, y)$ return a score increasing in $s_j$ when there is agreement, and decreasing in $s_j$ when there is disagreement (for positive $\theta_j$):
\begin{enumerate}
\item Treat $s$ as a weight: $\theta_{jy} \tau_j s_j$. 
\item Thresholded: $\theta_{jy} \tau_j \max(s_j - \pi_j, 0)$
\item Thresholded Sigmoid: $\theta_{jy} \tau_j \text{sigmoid}(s_j - \pi_j)$
\item Logit (Global conditioning on Beta): $\theta_{jy} \tau_j \log \frac{s}{1-s}$

\item Half clipped Gaussian instead of Beta with the Gaussian mean at '1' when $\tau_j$ and $y$ agree, and at 0 when they disagree.  The only learned parameter is then the variance of the Gaussian which we represent as $\pi_{jy}$.  


\begin{multline}
P(s_{ij} | y, l_{ij}, k_j) = 
\begin{cases}
    \textit{HG}(1 - s_{ij};\text{scale}=\theta _{jy0})  & \text{if $k_j=y ~\& ~l_{ij} \ne 0$,} \\
    \textit{HG}(s_{ij};\text{scale}=\theta _{jy1})  & \text{if $k_j \ne y ~\& ~l_{ij} \ne 0$,} \\
    1 & \text{otherwise.}
\end{cases}
\end{multline}
\end{enumerate}
In Table~\ref{tab:contPots} we compare accuracy with these alternative forms of potentials.   We appreciate the difficulty of choosing the right continuous potentials by observing  that for most of these other choices the accuracy is really poor.

\begin{table}[]
\begin{center}
\begin{tabular}{|l|l|l|l|l|l|l|}
\hline
$\psi(s_j,\tau_j,y)$ & \bf Spouse & \bf CDR   & \bf Sms   & \bf Dedup        & \bf Ionosphere & \bf Iris  \\ \hline
$\theta_{jy} \tau_j s_j$ & 0      & 0.49 & 0     & 0.26        & 0.96      & 0.87 \\ \hline
$\theta_{jy} \tau_j \max(s_j - \pi_j, 0)$  & 0      & 0.49 & 0     & 0            & 0.83      & 0.87 \\ \hline
$\theta_{jy} \tau_j \text{sigmoid}(s_j - \pi_j)$  & 0      & 0.49 & 0     & 0            & 0.80      & 0.87 \\ \hline
$\theta_{jy} \tau_j \log \frac{s}{1-s}$ & 0.46  & 0     & 0     & 0 & 0          & 0.44 \\ \hline
\bf \model  &  0.58  &  0.61   &  0.54     &  0.79 &  0.97          &  0.87 \\ \hline
\end{tabular}
\caption{Comparison of Continuous Potentials}
\label{tab:contPots}
\end{center}
\end{table}

\section{Example: Failure in spite of good initialization of the Snorkel Model}
\label{sec:majVsSnorkel}
Consider another dataset consisting of $n$ LFs that are just slightly better than random, that is their agreement with the true $y$ when they trigger is just (0.5+$\epsilon$)*100\%. 
%
A simple majority voting on the noisy LF labels will give nearly $100$\% accuracy when $n$ is large. On the other hand,  with the Snorkel model even with favorable initializations ($\theta_{j} = 1, \forall j)$, the final values of the parameters on covergence is often poor. In most training runs, the accuracy dropped to 0 after a few training iterations.  Our decoupled model is able to perform well in this case without accuracy guides.


\begin{figure}[ht]
\centering
\includegraphics[width=0.4\textwidth]{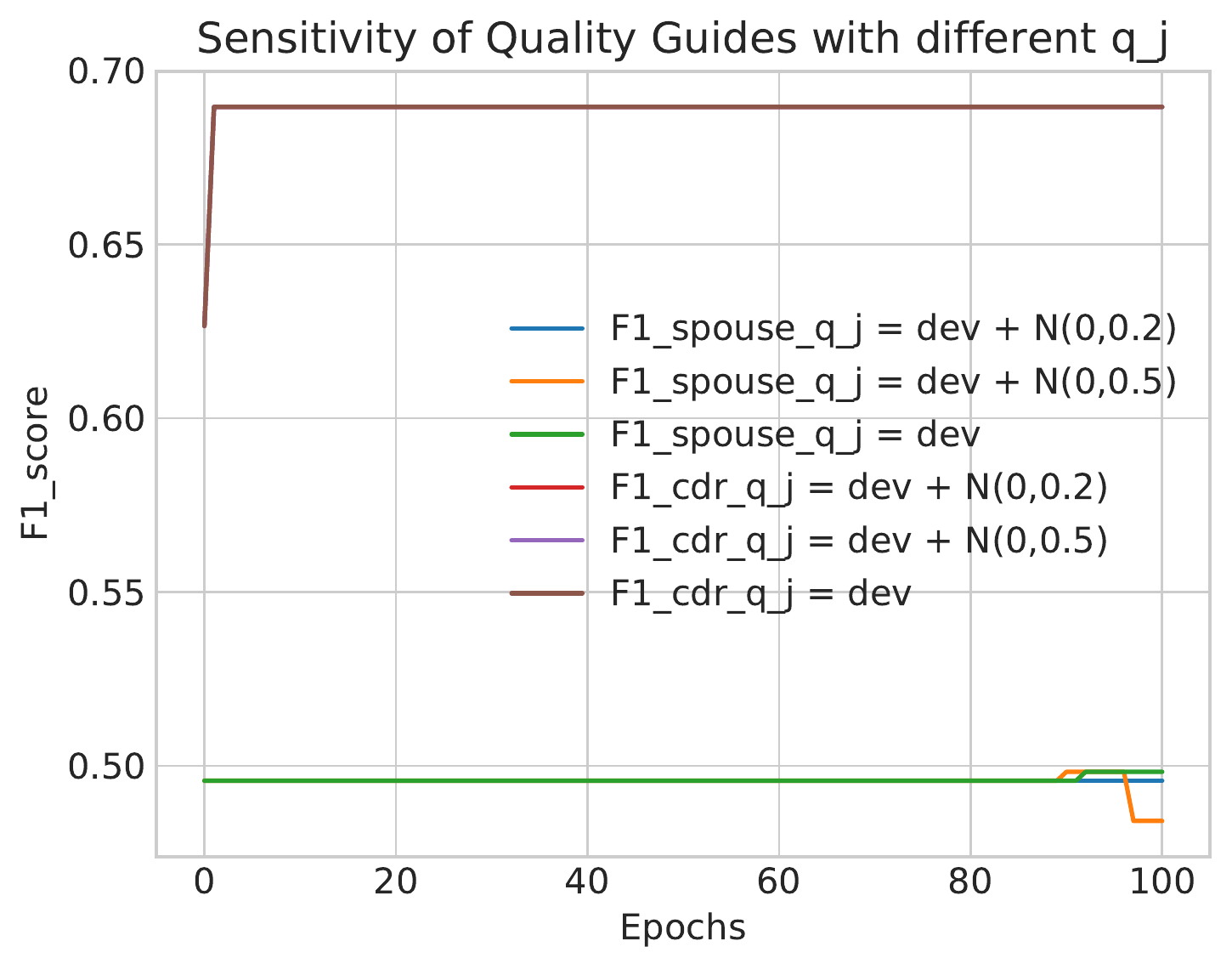}
\caption{F1 with increasing training epochs for three settings of $q_j^c$: (1) where $q_j^c=0.9$ for all LFs, (2) where $q_j^c=0.8$ for all LFs, and (3) where $q_j^c=$ true accuracy of LF $\lambda_j$ + a Gaussian noise of variance 0.2.  We observe that across two different datasets, the accuracy remains the same and stable for all three quality guides.  Without the guides, we already saw in Figure~\ref{fig:prec} of the main paper how the F1 swings with training epochs. }
\label{fig:stable}
\end{figure}

\section{Results by training Discriminative Classifier}
\label{sec:disc}
Data Programming is hinged upon creation of probabilistic training labels for any discriminative model with a standard loss function.
In Table~\ref{tab:disc}, we report precision, recall and F1 scores on the test data by training a logistic regression classifier on labeled data created by Snorkel as well as our generative model \model. A takeaway of the following results is that, through the use of continuous LFs and quality guides, the discriminative model generalizes much better beyond the heuristics encoded in the LFs.

\begin{table}[!htb]
\begin{center}
\begin{tabular}{|l|l|l|l|l|l|l|l|l|l|}
\hline
          & \multicolumn{3}{|c|}{\textbf{Spouse}} & \multicolumn{3}{|c|}{\textbf{CDR}} & \multicolumn{3}{|c|}{\textbf{Sms}}  \\ \hline
          & P & R & F1 & P & R & F1 & P & R & F1 \\ \hline
Snorkel   & 0.30   & 0.66    & 0.41 & 0.41 & 0.98 & 0.58 & 0.20 & 0.96     & 0.34  \\ \hline
\model & 0.65 & 0.47 &  0.55 & 0.58 & 0.86 & 0.69 & 0.52 & 0.66 & 0.58   \\ \hline
\end{tabular}
\caption{Discriminative Classifier trained on data labeled using the generative models and evaluated on held-out test data.  \label{tab:disc}}
\end{center}
\end{table}

\end{document}